\documentclass{article}

\usepackage{wrapfig}
\usepackage{CJKutf8} 
\usepackage{tabularx} 
\usepackage{longcat_style}

\usepackage[utf8]{inputenc} 
\usepackage[T1]{fontenc}    
\usepackage{newunicodechar}
\newunicodechar{，}{,}
\usepackage{hyperref}       
\usepackage{xcolor}
\usepackage{svg} 
\hypersetup{
    colorlinks=true,      
    linkcolor=blue,      
    urlcolor=blue,       
    citecolor=blue,      
    linkbordercolor=blue, 
    urlbordercolor=blue,
    citebordercolor=blue,
    pdfborderstyle={/S/U/W 1}, 
}
\usepackage{amsthm}
\newtheorem{definition}{Definition}
\usepackage{xurl}            
\usepackage{booktabs}       
\usepackage{amsfonts}       
\usepackage{nicefrac}       
\usepackage{microtype}      
\usepackage{lipsum}		
\usepackage{graphicx}
\usepackage{natbib}
\usepackage{doi}
\usepackage{amsmath}
\usepackage{amssymb} 
\usepackage{xspace}
\usepackage{enumitem}
\usepackage{multirow}
\usepackage{subcaption}
\usepackage{makecell}
\usepackage{hyperref, cleveref}
\setlist[itemize]{leftmargin=*}
\setlist[enumerate]{leftmargin=*}
\setlist[description]{leftmargin=*}
\newcommand{\topkattn}{Top-$k$ Attention\xspace}

\newcommand{\topkdecode}{Top-$k$ Decoding\xspace}

\newcommand{\topksft}{Top-$k$ SFT\xspace}

\newcommand{\vanilla}{Llama-3-8B-ProLong-512k-Instruct\xspace}

\newcommand{\base}{Llama-3-8B-ProLong-512k-Base\xspace}

\newcommand{\sft}{Llama-3-8B-ProLong-Instruct-512K-TopK-SFT\xspace}

\newcommand{\flashattn}{\textsc{FlashAttention}\xspace}

\usepackage{colortbl}
\definecolor{mygray}{gray}{.88}
\definecolor{mycyan}{cmyk}{.15,0,0,0}
\definecolor{mycyan2}{cmyk}{.85,0,0,0}

\definecolor{mygreen}{rgb}{0.19, 0.79, 0.02}
\definecolor{midnightgreen}{rgb}{0.0, 0.29, 0.33}

\title{A Preliminary Study on the Promises and Challenges of Native Top-$k$ Sparse Attention}

\author{
  Di Xiu\thanks{Equal contribution.} \quad
  Hongyin Tang\footnotemark[1] \quad
  Bolin Rong \\
  \bf{Lizhi Yan \quad Jingang Wang\thanks{Project Lead.} \quad Yifan Lu \quad Xunliang Cai} \\
  Meituan, Beijing, China. \\
  \texttt{\{xiudi, tanghongyin, rongbolin\}@meituan.com} \\
  \texttt{\{yanlizhi, wangjingang02, luyifan04, caixunliang\}@meituan.com}
}

\renewcommand{\headeright}{\raisebox{-0.2\height}{\includegraphics[height=1.8em]{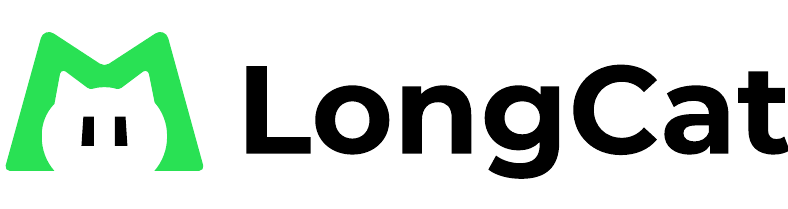}}}
\renewcommand{\shorttitle}{A Preliminary Study on the Promises and Challenges of Native Top-$k$ Sparse Attention}

\begin{document}
\maketitle

\begin{abstract}


Large Language Models (LLMs) are increasingly prevalent in the field of long-context modeling, however, their inference computational costs have become a critical bottleneck hindering the advancement of tasks such as agents and multimodal applications. Centering on the core issue of accelerating long-context inference, this report conducts a preliminary investigation into the effectiveness and theoretical mechanisms of the \topkattn mechanism during both the decoding and training phases.
First, we validate the effectiveness of exact \topkdecode through extensive experimentation. Experiments based on models such as Llama-3-8B-ProLong-Instruct-512K and Qwen3-32B demonstrate that retaining only the pivotal Keys with the highest similarity to the Query as the context window during the decoding stage not only significantly reduces computational overhead but also achieves performance comparable to, or even surpassing, full attention on downstream tasks such as HELMET and LongBench v2.
Second, we further explore the native \topkattn training strategy. By fine-tuning \base with the introduction of the \topkattn kernel during the Supervised Fine-Tuning (SFT) stage, our experiments confirm that ensuring the consistency between training and inference regarding \topkattn operations facilitates the further unlocking of \topkdecode's potential, thereby significantly enhancing model performance.
Furthermore, considering the high computational complexity of exact \topkattn, we investigate the impact of approximate Top-$k$ algorithm precision on downstream tasks. Our research confirms a positive correlation between downstream task performance and approximation fidelity, and we provide statistical evaluations of the Lightning Indexer's precision within the DeepSeek-V3.2-Exp model.
Finally, this report provides a theoretical interpretation from the perspective of Entropy. Experimental observations indicate that models subjected to \topkattn SFT exhibit a distinct phenomenon of entropy reduction in downstream tasks. This validates the hypothesis that low-entropy states are better adapted to \topkdecode, providing a robust theoretical foundation for the effectiveness of sparse attention mechanisms.

\end{abstract}


\newpage
\section{Exact \topkdecode Suffices}

The long-context modeling capability of Large Language Models (LLMs) increasingly constrains the advancement of applications such as agents and multimodality, emerging as a critical technical bottleneck requiring urgent resolution  \citep{liu2025comprehensive}. Consequently, accelerating long-context inference has garnered significant attention as a pivotal research topic  \citep{chen2024magicpiglshsamplingefficient, tang2024quest, xiao2024infllm, tang2024ltri, cai2025rkv}.

\begin{figure}[htbp]
    \centering
    \begin{subfigure}[b]{0.64\linewidth} 
        \centering
        \includegraphics[width=\linewidth]{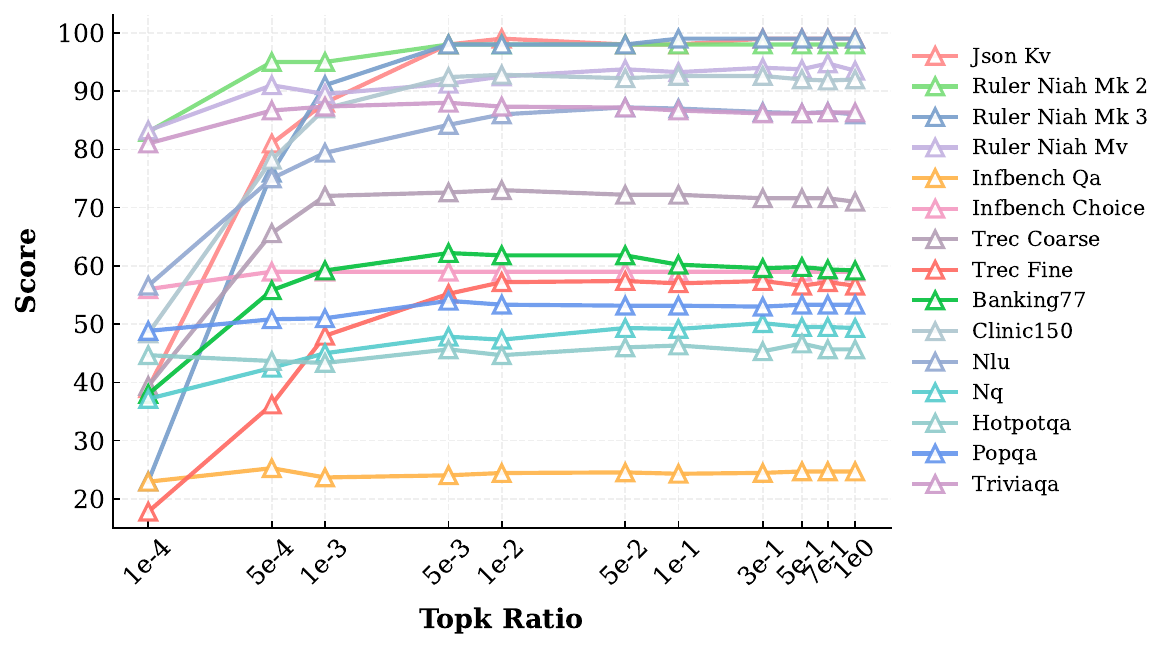}
        \caption{\small \vanilla performance on HELMET-128K benchmark.}
        \label{fig:prolong_helmet}
    \end{subfigure}

    \vspace{0.5cm} 

    \begin{subfigure}[b]{0.64\linewidth}
        \centering
        \includegraphics[width=\linewidth]{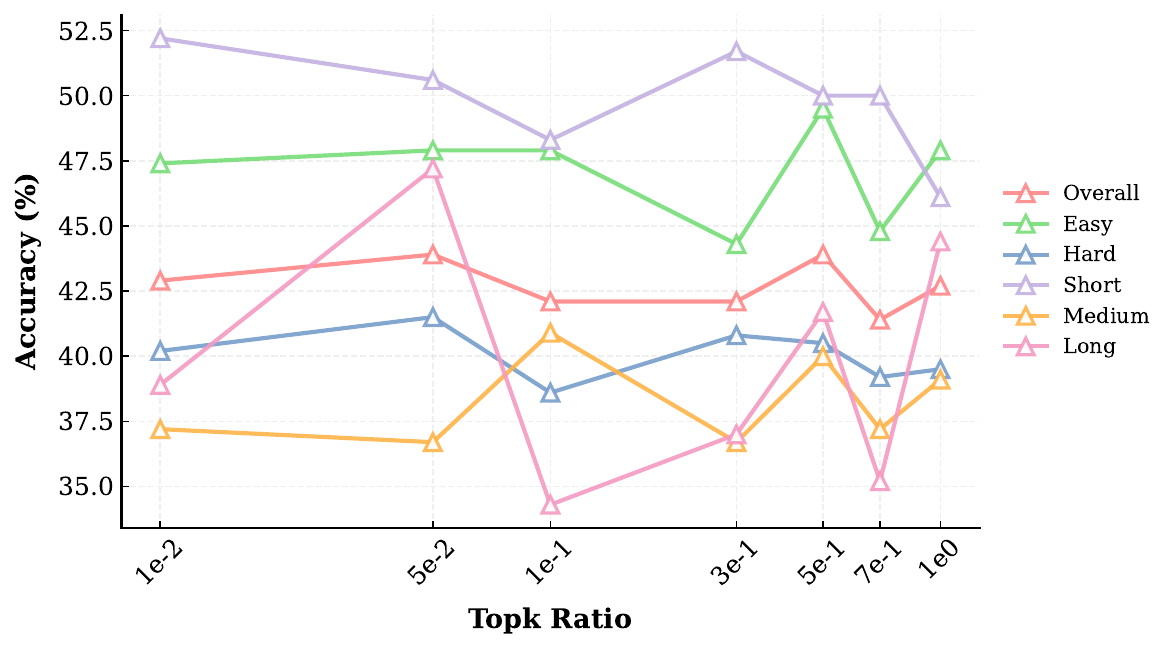}
        \caption{\small Qwen3-32B non-thinking mode performance on LongBench v2.}
        \label{fig:qwen3_32b_longbenchv2}
    \end{subfigure}

    \caption{Performance of \topkdecode on different benchmarks.} 
    \label{fig:combined_results}
\end{figure}

We evaluated exact \topkdecode \citep{synk2025exploiting} with \vanilla \citep{gao2025prolong} and Qwen3-32B \citep{yang2025qwen3} across benchmarks such as HELMET \citep{yen2024helmet}, MATH500 \citep{math500}, GSM8K \citep{cobbe2021gsm8k}, LongBench v2 \citep{bai2025longbenchv2} and AIME24 \citep{AIME24}. Empirical results in Figure \ref{fig:combined_results} demonstrates that \topkdecode rivals or outperforms full attention even at low Top-$k$ ratios.

\begin{definition}[\textbf{Top-$k$ Ratio}]
\label{def:topk_ratio}
Let $\mathcal{K} = \{k_1, k_2, \dots, k_N\}$ denote the set of all key tokens in the context, where $N$ is the total context length. During the decoding phase for a given query token, let $\mathcal{K}_{top} \subset \mathcal{K}$ represent the subset of selected key tokens consisting of the $\mathcal{W}$ most prominent tokens, such that $|\mathcal{K}_{top}| = \mathcal{W}$. During the attention calculation, the query token interacts exclusively with keys in $\mathcal{K}_{top}$, ignoring the rest.

The \textbf{Top-$k$ Ratio}, denoted as $\rho$, is defined as the proportion of the selected critical tokens to the total context tokens:
\begin{equation}
    \rho = \frac{\mathcal{W}}{N} = \frac{|\mathcal{K}_{top}|}{|\mathcal{K}|}
\end{equation}
where $\rho \in (0, 1]$. A lower $\rho$ indicates a sparser attention computation, while $\rho=1$ corresponds to full attention.\looseness=-1
\end{definition}
\vspace{-0.5em}

\newpage
\section{Native \topkattn Training Promotes \topkdecode}





Motivated by the promising observation that \topkattn achieves performance comparable to full attention even at low Top-$k$ ratios, we noted that the \vanilla model used in our initial experiments was originally trained in the full attention way during both the long-context continued training and supervised fine-tuning of the Meta-Llama-3-8B-Instruct model \citep{dubey2024llama}. This prompted further investigation: would a model natively trained with \topkattn yield superior results when applying \topkdecode?

To this end, we utilized the \base \footnote{\url{https://huggingface.co/princeton-nlp/Llama-3-8B-ProLong-512k-Base/tree/main}} model (the checkpoint preceding the SFT stage of the Instruct model) to conduct subsequent SFT. We employed the same SFT corpus as the original Instruct model but integrated the \topkattn training kernel, resulting in the \sft model. Subsequently, we evaluated the relative performance of the \topksft model against the original Instruct model on downstream tasks using the 8K variant of HELMET benchmark.

Specifically, at the training kernel level, we wrapped the \flashattn \citep{tridao2022flashattn} kernel. By pre-calculating the exact Top-$k$ indices and scores and passing them into the kernel to update the corresponding masks, we implemented variable-length \topkattn training at Top-$k$ ratio of 1\%.

Empirical results demonstrate that the \sft model exhibits significant performance improvements over the original model on downstream tasks. Given that this adaptation was achieved using a limited number of tokens solely during the SFT phase, we anticipate that incorporating \topkattn during the continued training phase with a larger token corpus would yield even greater performance gains.

\begin{figure}[htbp]
\centerline{\includegraphics[width=\linewidth]{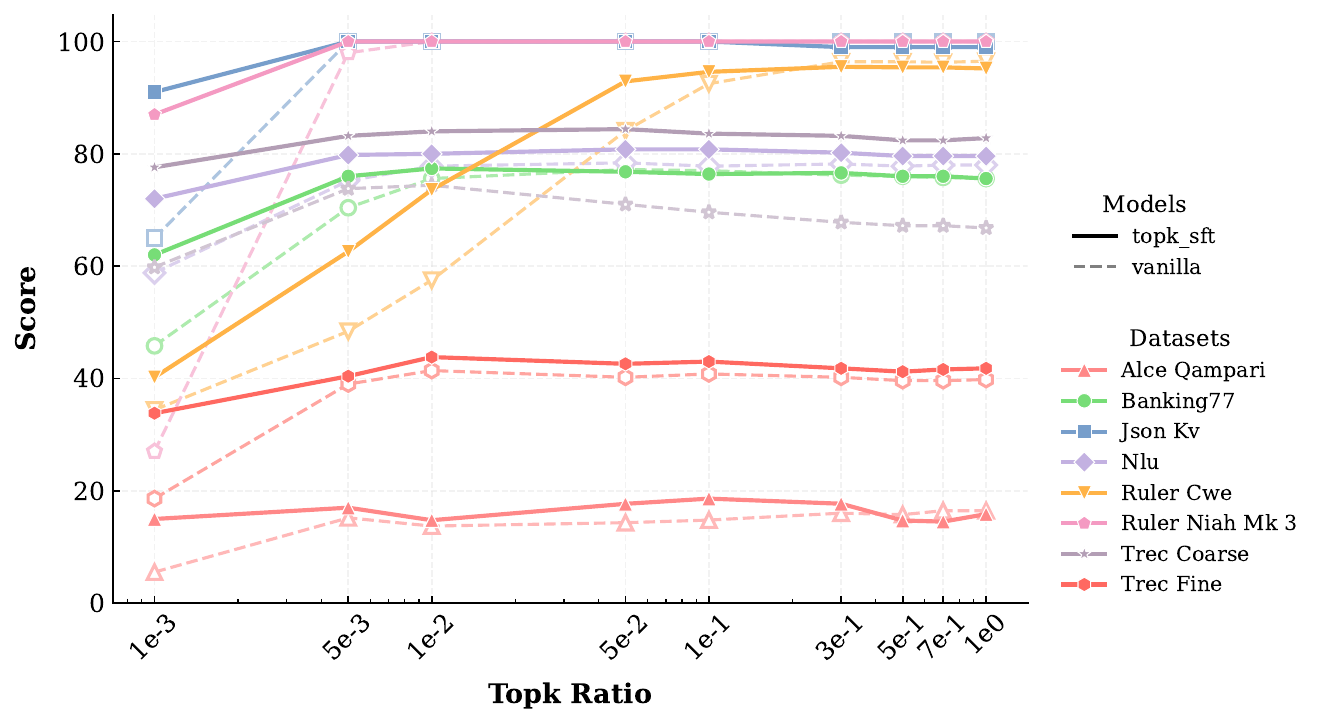}}
\caption{\small
\topkdecode performance on the 8K variant of HELMET benchmark between \vanilla and \topksft.
}
\label{fig:prolong_sft_advantage}
\end{figure}

\newpage
\section{Precision Impact for Approximate \topkdecode}

In the preceding sections, the approaches to \topkattn we discussed---whether applied directly during the decoding phase or utilized following Supervised Fine-Tuning ---share a common characteristic: they all fall within the scope of exact \topkattn.

It is important to note that computing exact Top-$k$ attention entails high computational complexity. A naive calculation of attention scores costs $\mathcal{O}(N^2)$ time and space complexity. Furthermore, to mitigate Out-Of-Memory (OOM) issues, frequent CPU-GPU offloading may be necessitated during the decoding phase, which further impedes the algorithm's execution speed.

One of the core design principles of \flashattn---the \textit{de facto} standard for modern full attention---is to avoid materializing logits by performing attention computations in a tiling fashion. This design presents a challenge for compatibility with exact Top-$k$ attention. As Tri Dao, the author of \flashattn, explicitly stated: ``If you want attn scores it's better to just use the standard implementation.''\footnote{\url{https://github.com/Dao-AILab/flash-attention/issues/1824}}

Consequently, the practical implementation of Top-$k$ attention may require the adoption of corresponding approximation algorithms, such as utilizing Approximate Nearest Neighbor (ANN) \citep{chen2018sptag, douze2024faiss} algorithms to rapidly construct approximately accurate indices. To formally assess the fidelity of such approximations compared to the exact baseline, we introduce the following metric:

\begin{definition}[\textbf{Retrieval Precision}]
\label{def:retrieval_precision}
Let $\mathcal{K} = \{k_1, k_2, \dots, k_N\}$ denote the set of all key tokens in the context, where $N$ is the total context length. Given a specific context window size $\mathcal{W}$, let $\mathcal{K}_{top} \subset \mathcal{K}$ denote the set of tokens corresponding to the exact top-$\mathcal{W}$ attention scores, and let $\mathcal{K}_{approx} \subset \mathcal{K}$ denote the set of tokens actually retrieved by an approximate method, such that $|\mathcal{K}_{top}| = |\mathcal{K}_{approx}| = \mathcal{W}$. 

The \textbf{Retrieval Precision}, denoted as $p$, is defined as the overlap ratio between the retrieved and ground-truth token sets:
\begin{equation}
    p = \frac{|\mathcal{K}_{approx} \cap \mathcal{K}_{top}|}{|\mathcal{K}_{approx}|} = \frac{|\mathcal{K}_{approx} \cap \mathcal{K}_{top}|}{\mathcal{W}}
\end{equation}
where $p \in [0, 1]$. While the Top-$k$ Ratio $\rho$ determines the \textit{sparsity} of the computation, the precision $p$ measures the \textit{fidelity} of the retrieval. A higher $p$ implies that the approximate method preserves more of the critical information identified by full attention.
\end{definition}

\begin{figure}[htbp]
\centerline{\includegraphics[width=\linewidth]{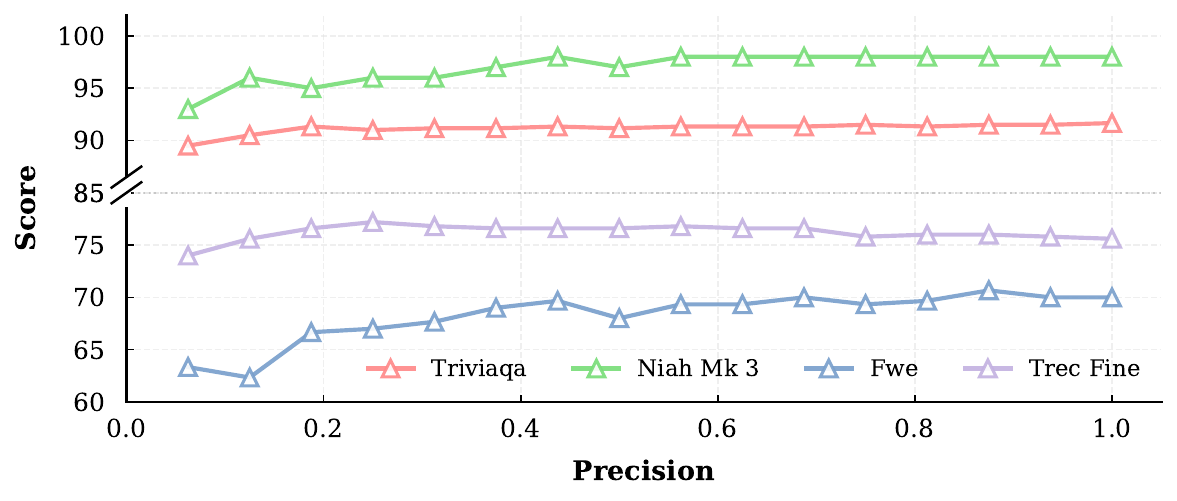}}
\caption{\small
\topkdecode precision-performance curve on the 128K variant of HELMET benchmark for \vanilla with context window size being 2048.
}
\label{fig:precision_performance_curve}
\end{figure}

An intuitive hypothesis is that given a specific context window size $\mathcal{W}$ and an approximate Top-$k$ retrieval precision $p$, performance on downstream tasks will improve as $p$ increases. To validate this intuition, we evaluated the performance of the Llama-3-8B-ProLong-Instruct-512K model on the 128K variant of HELMET benchmark across varying levels of retrieval precision $p$. Specifically, for a given dataset $\mathcal{D}$, we maintained the use of \flashattn-based Full Attention during the prefilling phase. During the decoding phase, we fixed the context window length $\mathcal{W}$ for each attention head at 2048 and adjusted the number of tokens within this window that correspond to the exact Top-$k$ indices to $2048 \times p$. We then calculated the corresponding downstream task accuracy. The results illustrated in Figure \ref{fig:precision_performance_curve} preliminarily corroborate our hypothesis that downstream task performance improves as precision $p$ increases until saturation.

The Lightning Indexer employed by the contemporaneous DeepSeek-V3.2-Exp model \citep{deepseekai2024deepseekv32}, despite theoretically retaining $\mathcal{O}(N^2)$ complexity, effectively reduces the coefficient of computational complexity through a design featuring fewer attention heads, reduced embedding dimensions, and FP8 quantization. Thus, it serves as an approximation method for exact Top-$k$ attention. We deployed DeepSeek-V3.2-Exp using SGLang \citep{zhang2024sglang}, applying the recent official patch regarding RoPE \citep{su2024roformer} positional embeddings revisions\footnote{\url{https://github.com/sgl-project/sglang/pull/13495}}, and evaluated the precision of the Lightning Indexer on the HELMET benchmark.

\begin{figure}[htbp]
    \centering
    \includegraphics[width=0.48\linewidth]{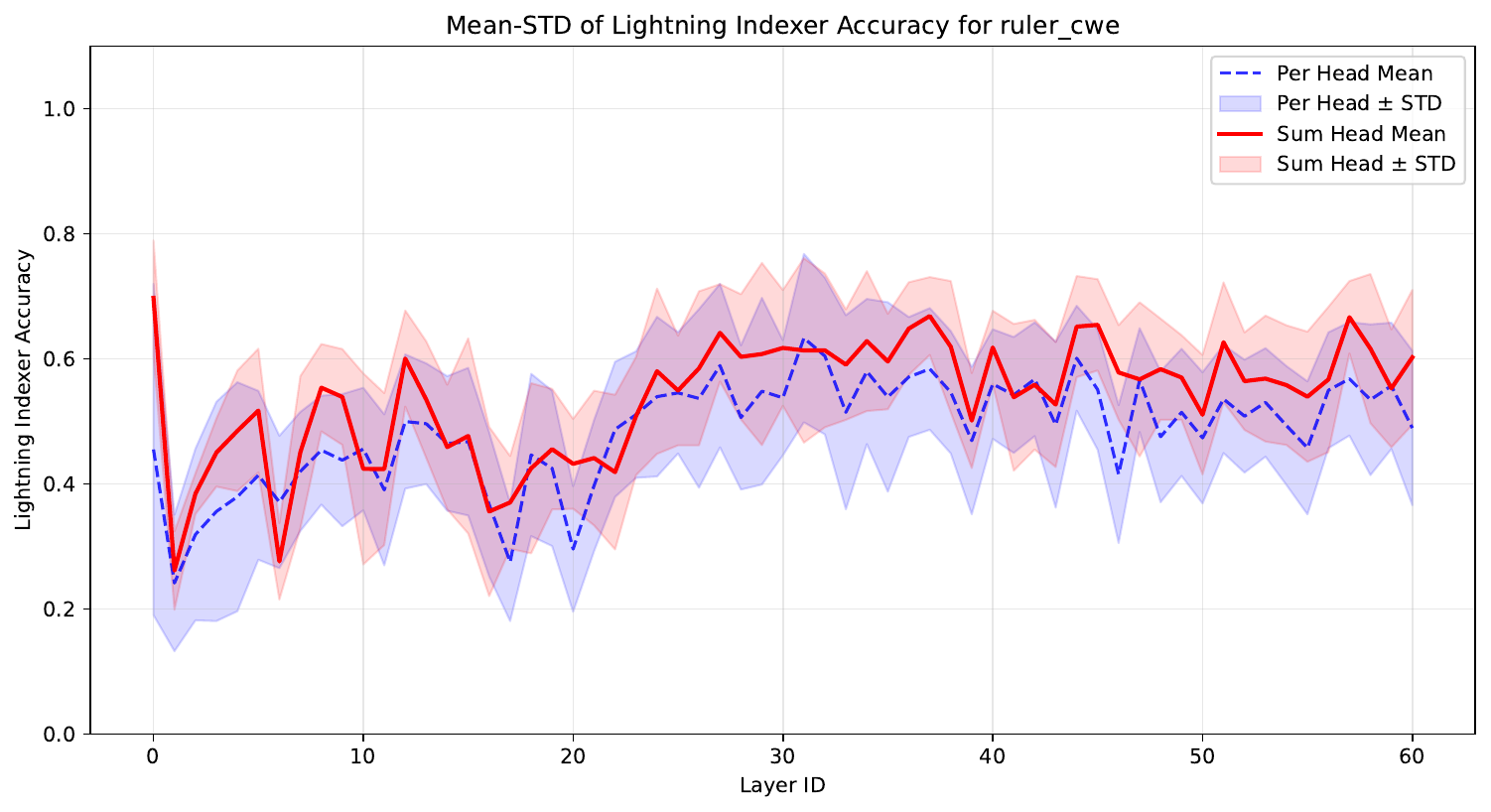} \hfill
    \includegraphics[width=0.48\linewidth]{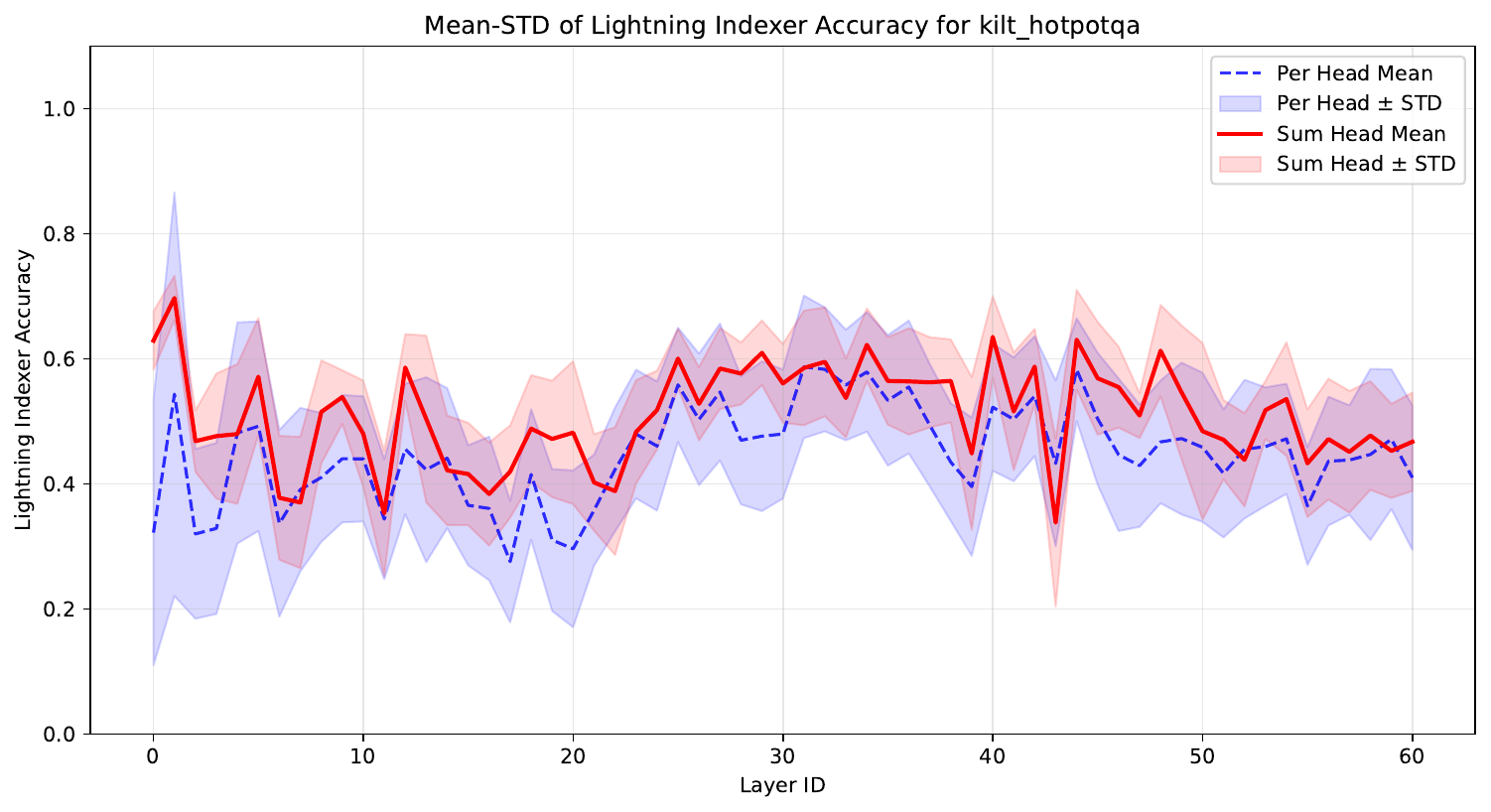} \\ 

    \vspace{0.2cm} 

    \includegraphics[width=0.48\linewidth]{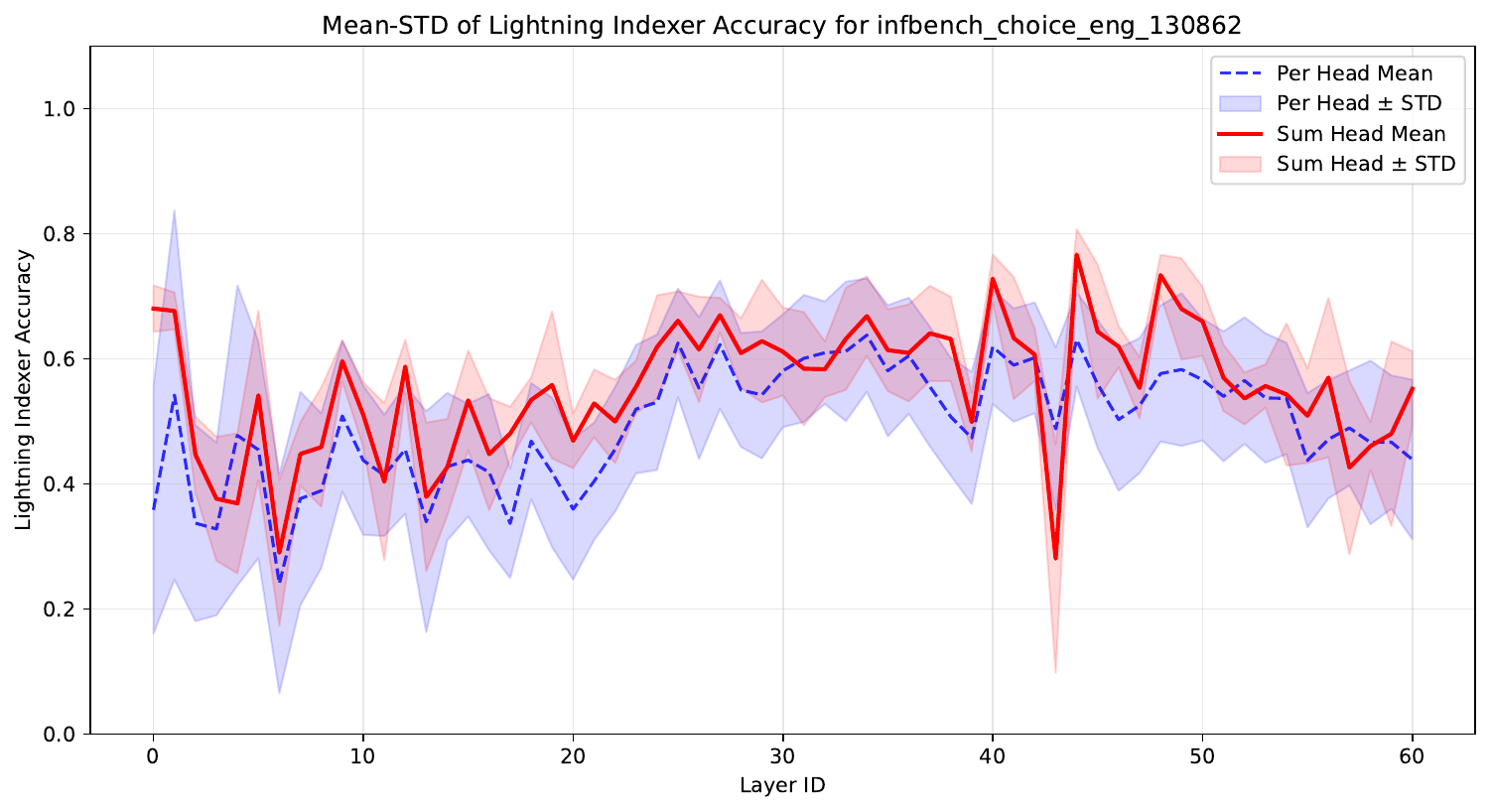} \hfill
    \includegraphics[width=0.48\linewidth]{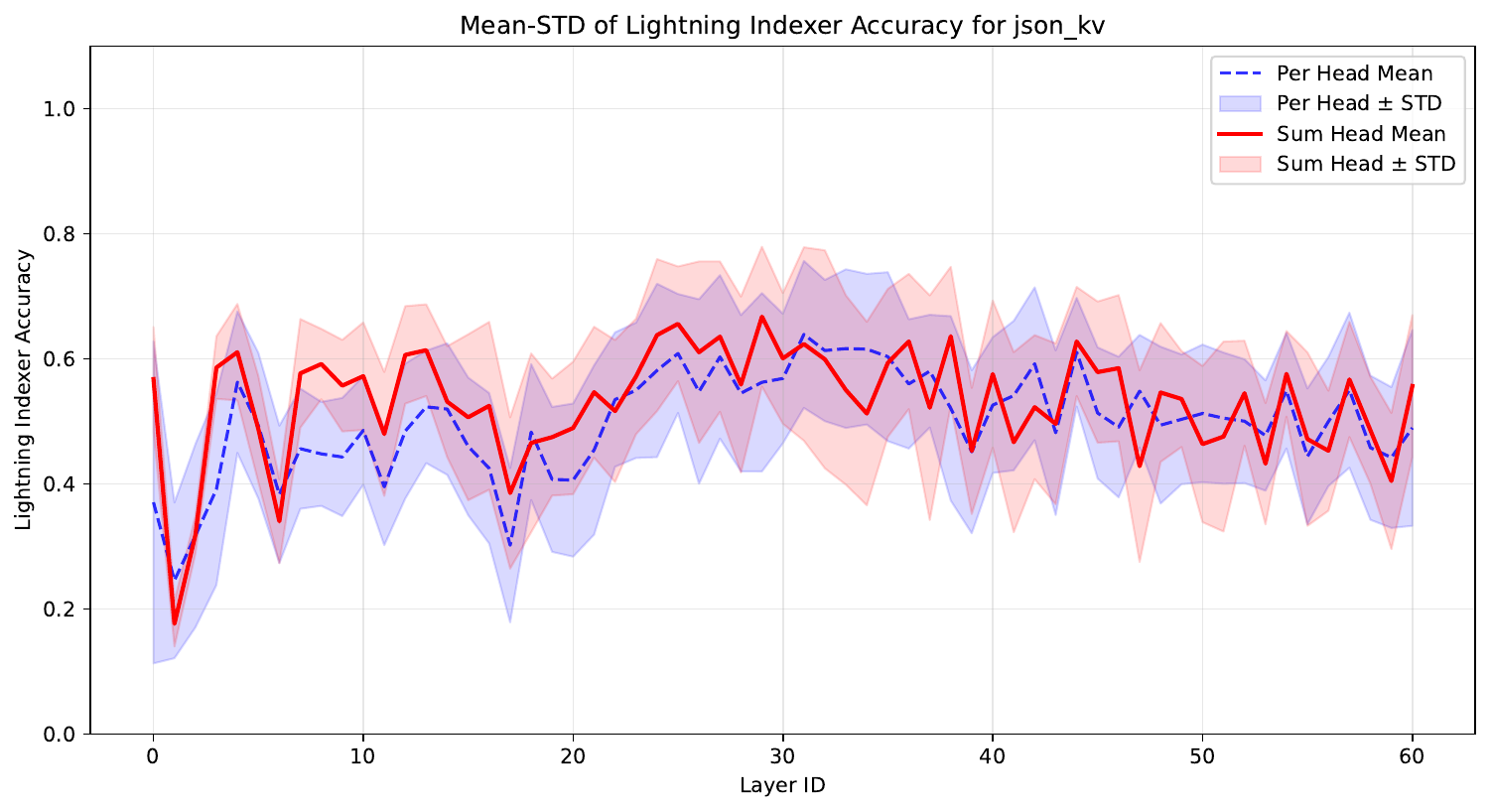}

    \caption{\small
    Visualization of layer-wise statistics for four representative datasets.
    Top-left: Dataset A; Top-right: Dataset B; Bottom-left: Dataset C; Bottom-right: Dataset D.
    These plots demonstrate the variance in feature distributions across different layers.
    }
    \label{fig:lightning_indexer_precision}
\end{figure}

The specific evaluation methodology was as follows: for a given dataset, ten samples were selected to generate statistical information. For a specific layer, the precision of the Lightning Indexer was calculated as the average overlap ratio between the exact Top-$k$ indices of the main model's various attention heads and the Lightning Indexer's Top-$k$ indices, across all decoding steps for all test samples.

Although the precision of the Lightning Indexer module in DeepSeek-V3.2-Exp averages approximately 60\% across different datasets in the 128K variant of HELMET benchmark, the model's end-to-end performance on the benchmark remains superior, bolstered by its massive parameter scale.

An interesting observation is that while previous exact \topkattn mechanisms operated on a \textit{per-head} basis, the Lightning Indexer in DeepSeek-V3.2-Exp is designed such that different attention heads share the same set of KV tokens. The results obtained from two different methods of calculating Lightning Indexer accuracy align closely, indicating that DeepSeek-V3.2-Exp achieves an approximation of exact \topkattn via the Multi-Query Attention (MQA) \citep{shazeer2019fast} mode, a more memory-efficient approach.

\newpage
\section{Entropy Perspective for \topkdecode}

Experiments on \topkdecode demonstrate that utilizing a reduced context window during the decoding phase yields performance comparable to that of full attention. Drawing on the relationship between Top-$k$ and Top-$p$ \citep{lin2025twilight} sampling, entropy serves as a valuable analytical metric. Specifically, we hypothesize that \topkdecode is particularly well-suited for tasks characterized by lower entropy. A natural corollary of this hypothesis is that \sft should exhibit a reduction in entropy on downstream tasks compared to the \vanilla model.

\begin{figure}[htbp]
\centerline{\includegraphics[width=\linewidth]{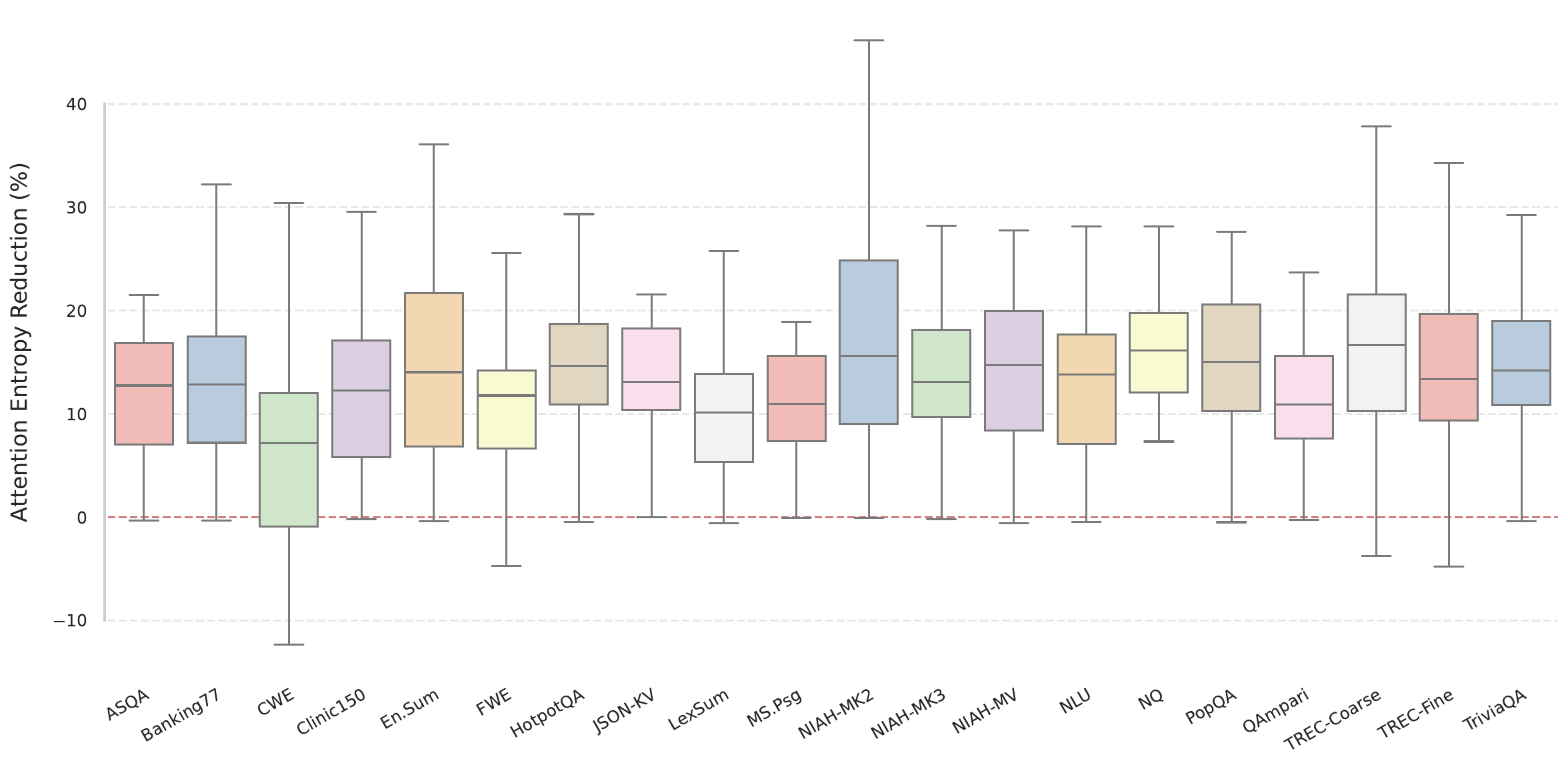}}
\caption{\small
  Attention entropy reduction (\%) of \sft compared to \vanilla on the 8K variant of HELMET benchmark.
}
\label{fig:attn_entropy_reduction}
\end{figure}

To validate this hypothesis, we compared the entropy values derived from the two aforementioned models across various tasks within the 8K variant of HELMET benchmark, employing distinct statistical strategies. The experimental aggregation space encompasses the following dimensions: the scope of attention heads (all attention heads vs.\ retrieval heads \citep{wu2024retrieval}); the aggregation method for the entropy of the attention heads of interest (mean, min, max, or median); the inference phase (prefilling or decoding); and the number of tokens included in the computation (1, 5, or 10). The obtained experimental results largely align with the expectation that the \sft model exhibits lower entropy, as depicted in Figure \ref{fig:attn_entropy_reduction}.

\newpage
\section{Conclusion}


\noindent This report investigates the application potential, optimization strategies, and underlying mechanisms of \topkattn in long-context reasoning for Large Language Models (LLMs), yielding the following key conclusions:

\begin{enumerate}
    \item Adopting exact \topkattn during the decoding phase is a highly promising acceleration strategy. Even at low Top-$k$ ratios, this method maintains high performance comparable to full attention across various long-context and logical reasoning benchmarks, effectively mitigating the computational bottlenecks associated with long-context reasoning.

    \item Incorporating native \topkattn training significantly enhances model performance. By utilizing \topkattn operators during the Supervised Fine-Tuning phase, the model adapts more effectively to the sparse attention patterns characteristic of \topkdecode during inference.

    \item Addressing the complexity of exact \topkattn, we demonstrated a positive correlation between the precision of approximate retrieval algorithms and model performance. Analysis of the DeepSeek-V3.2-Exp model indicates that despite the limited precision of approximate algorithms (e.g., Lightning Indexer, $\sim$60\%), the integration of massive parameters enables superior end-to-end performance. These findings provide valuable insights for the future design of low-complexity approximate Top-$k$ operators.

    \item From an information-theoretic perspective, \topksft effectively reduces the model's entropy when processing downstream tasks, corroborating the inherent advantages of \topkdecode in low-entropy task environments.
\end{enumerate}

\bibliographystyle{unsrtnat}
\bibliography{references}

\end{document}